\documentclass[sigconf]{acmart}

\copyrightyear{2023}
\acmYear{2023}
\setcopyright{acmlicensed}
\acmConference[MM '23] {Proceedings of the 31st ACM International Conference on Multimedia}{October 29--November 3, 2023}{Ottawa, ON, Canada.}
\acmBooktitle{Proceedings of the 31st ACM International Conference on Multimedia (MM '23), October 29--November 3, 2023, Ottawa, ON, Canada}
\acmPrice{15.00}
\acmISBN{979-8-4007-0108-5/23/10}
\acmDOI{10.1145/3581783.3612105}

\AtBeginDocument{%
  \providecommand\BibTeX{{%
    \normalfont B\kern-0.5em{\scshape i\kern-0.25em b}\kern-0.8em\TeX}}}

\usepackage{multirow}
\usepackage{nicematrix}
\usepackage{graphicx}
\usepackage{tabularx}

\usepackage{marvosym}
\usepackage[capitalize]{cleveref}

\begin{document}
\title{Ground-to-Aerial Person Search: Benchmark Dataset and Approach}



\author[]{Shizhou Zhang}
\email{szzhang@nwpu.edu.cn}
\affiliation{%
  \institution{Northwestern Polytechnical University}
  \city{Xi'an}
  \state{Shaanxi}
  \country{China}
  \postcode{710072}
}
\author[]{Qingchun Yang}
\email{yqc123@mail.nwpu.edu.cn}
\affiliation{%
  \institution{Northwestern Polytechnical University}
  \city{Xi'an}
  \state{Shaanxi}
  \country{China}
  \postcode{710072}
}

\author[]{De Cheng}
\authornote{Corresponding author}
\email{dcheng@xidian.edu.cn}
\affiliation{%
  \institution{Xidian University}
  \city{Xi'an}
  \state{Shaanxi}
  \country{China}
  \postcode{710126}
}

\author[]{Yinghui Xing}
\email{xyh_7491@nwpu.edu.cn}
\affiliation{%
  \institution{Northwestern Polytechnical University}
  \city{Xi'an}
  \state{Shaanxi}
  \country{China}
  \postcode{710072}
}

\author[]{Guoqiang Liang}
\email{gqliang@nwpu.edu.cn}
\affiliation{%
  \institution{Northwestern Polytechnical University}
  \city{Xi'an}
  \state{Shaanxi}
  \country{China}
  \postcode{710072}
}

\author{Peng Wang}
\email{peng.wang@nwpu.edu.cn}
\affiliation{%
  \institution{Northwestern Polytechnical University}
  \city{Xi'an}
  \state{Shaanxi}
  \country{China}
  \postcode{710072}
}

\author{Yanning Zhang}
\email{ynzhang@nwpu.edu.cn}
\affiliation{%
  \institution{Northwestern Polytechnical University}
  \city{Xi'an}
  \state{Shaanxi}
  \country{China}
  \postcode{710072}
}


\renewcommand{\shortauthors}{Shizhou Zhang et al.}
\begin{abstract}

In this work, we construct a large-scale dataset for Ground-to-Aerial Person Search, named G2APS, which contains 31,770 images of 260,559 annotated bounding boxes for 2,644 identities appearing in both of the UAVs and ground surveillance cameras. To our knowledge, this is the first dataset for cross-platform intelligent surveillance applications, where the UAVs could work as a powerful complement for the ground surveillance cameras. To more realistically simulate the actual cross-platform Ground-to-Aerial surveillance scenarios, the surveillance cameras are fixed about 2 meters above the ground, while the UAVs capture videos of persons at different location, with a variety of view-angles, flight attitudes and flight modes. Therefore, the dataset has the following unique characteristics: 1) drastic view-angle changes between query and gallery person images from cross-platform cameras; 2) diverse resolutions, poses and views of the person images under 9 rich real-world scenarios. On basis of the G2APS benchmark dataset, we demonstrate detailed analysis about current two-step and end-to-end person search methods, and further propose a simple yet effective knowledge distillation scheme on the head of the ReID network, which achieves state-of-the-art performances on both of the G2APS and the previous two public person search datasets, i.e., PRW and CUHK-SYSU. The dataset and source code available on \url{https://github.com/yqc123456/HKD_for_person_search}.

\end{abstract}




\maketitle



\begin{figure}[htbp] 
   \centering 
   \includegraphics[width=0.47\textwidth,height=0.20\textwidth]{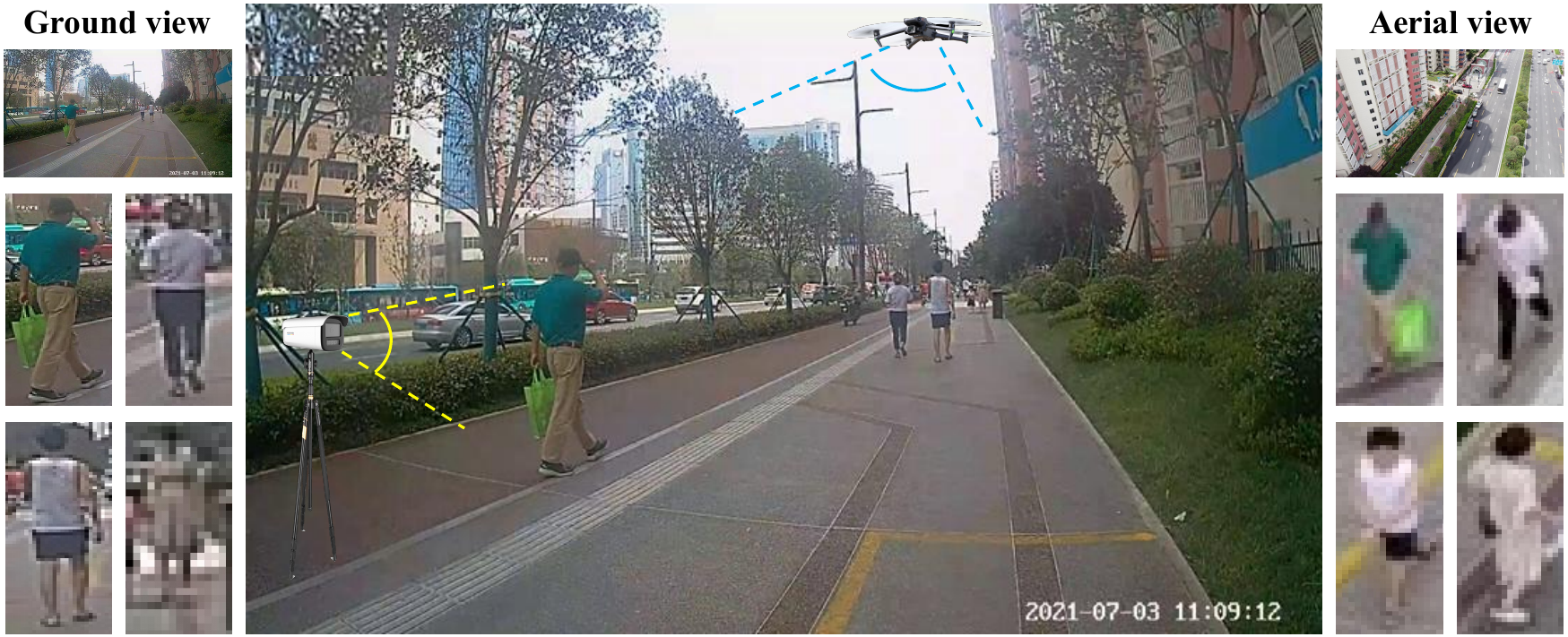} 
   \caption{A real world scenario for capturing images of our cross-platform Ground-to-Aerial person search dataset.} 
   \label{fig:concept} 

\end{figure}

\section{Introduction}
Recently, the Unmanned Aerial Vehicles (UAV)-based vision applications have drawn increasing attentions from both of the industry and academic sectors, as their practical application values in the real-world scenarios.
Existing UAV-related research and datasets mainly focus on the tasks of object detection~\cite{zhu2018visdrone,mittal2020deep,liu2020uav}, object tracking~\cite{du2019visdrone,liu2022multi,chen2017real}, action recognition~\cite{othman2021challenges,li2021uav,perera2019drone}, etc. However, the UAV-based person ReID and person search have rarely been studied.
The main reason is the lack of corresponding cross-platform Ground-to-Aerial dataset, which will take a large amount of human efforts for UAV flying, video capture and data annotations.

Specially for these cross-platform person identity annotation, it needs to match the same identity across UAV and ground cameras, which takes much more effort than identity annotation under the same video capture platform. Existing person ReID and person search datasets~\cite{xiao2017joint,zheng2017person,zheng2015scalable,zheng2017unlabeled,wei2018person} are collected by fixed surveillance cameras under the single video-capture platform. 
Although there appears one person ReID dataset in aerial imagery~\cite{zhang2020person} recently, images/videos in them are only captured and annotated under the single UAV platform. 
In contrast, the cross-platform Ground-to-Aerial surveillance system is much more advanced and practical. 
Suppose that if we want to find a suspect/person of interest in rural areas where there is no ground surveillance cameras deployed.
And the only available information is a query image captured by a ground camera.  
One feasible solution is to search the person of interest with the help of a UAV mounted with a camera.
In such scenarios, it is essential to develop the technique of ground-to-aerial person search which will suffer from severer intra-class object changes due to the large view-angle, image resolution and quality differences in cross video-capture platforms.  

In this paper, we construct a large-scale Ground-to-Aerial person search dataset for the cross-platform Ground-to-Aerial intelligent surveillance applications, named Ground-to-Aerial Person Search (G2APS). The G2APS dataset consists of 31,770 images of 260,559 annotated bounding boxes, of which 199,696 bounding boxes are labeled with 2,644 identities. Note that, these 2,644 identities appear in both of the UAV and ground surveillance cameras. The 60,863 person bounding boxes are labeled with -1, where the corresponding persons only appear in one single device. On average, there are about 75 bounding boxes for each identity in the G2APS dataset, which is much more than PRW~\cite{zheng2017person} and CUHK-SYSU~\cite{xiao2017joint} datasets.

The images of each person instance are captured by cameras of a DJI consumer UAV and a ground surveillance camera. 
In order to more realistically simulate the cross-platform Ground-to-Aerial surveillance scenarios, the ground surveillance cameras are fixed about 2.0 meters above the ground, while the UAV captures videos of persons at different location, with a variety of view-angles, flight attitudes and flight modes. 
Specifically, the dataset is collected from nine different locations, including primary school campus, university campus, subway station entrance, tourist sites, crossroads, sidewalk  etc. 
The flight attitudes varies from 20 meters to 60 meters, and the flight mode includes hovering, cruising and rotating, which makes the dataset contain rich perspectives. 
As shown in Figure~\ref{fig:concept}, the task of cross-platform Ground-to-Aerial person search is typically more challenging than the traditional single-platform counterpart, where the images are captured only by the fixed ground surveillance cameras, as the persons in the Ground-to-Aerial surveillance scenarios are featured in large view-point and pose variations, and also wider range of image resolution.

To deeply analyze existing person search methods on the newly proposed cross-platform Ground-to-Aerial person search tasks, we conduct extensive experiment comparisons including current representative two-step and end-to-end person search approaches. 
Experiment results demonstrate that the end-to-end person search methods always obtains inferior performances than that of the two-step methods, as shown in ~\autoref{tab:end_to_end_and_two_step}. 
The main reason is the conflicting optimization objectives between the position regression, foreground-background classification and fine-grained person re-identification loss, where the position regression aims to learn boundary profile features of the target, while the ReID loss aims to learn fine-grained discriminative feature representations. 

Inspired by the above analysis, we propose a simple yet efficient knowledge distillation scheme on the head of ReID network, without introducing any extra computation cost during model inference, while with only a very small amount of extra computation cost during model training. 
Specifically, the proposed ReID distillation branch is constructed on top of the backbone network features, which will have few interference on the object detection tasks. 
Thus, it helps to improve the ReID performance without deteriorating the detection performances.

To summarize, the main contributions of this paper are as follows:
\begin{itemize}
\item We are the first to construct a large-scale Ground-to-Aerial Person Search (G2APS) benchmark dataset for the cross-platform ground-to-aerial intelligent surveillance applications. The G2APS dataset consists of 31,770 images of 260,559 annotated bounding boxes, for 2,644 identities appearing in both of the UAV and ground surveillance cameras. 
\item This dataset has the following unique characteristics: 1) drastic view-angle changes between query and gallery person images from cross-platform cameras; 2) diverse resolutions, poses and views of the person images under nine rich real-world scenarios. 
\item  On top of the G2APS benchmark dataset, we give detailed analysis about current two-step and end-to-end person search methods, and further propose a simple yet effective knowledge distillation scheme on the ReID network head. The proposed method achieves state-of-the-art performances on both the G2APS and the previous public PRW and CUHK-SYSU datasets.
\end{itemize}

\graphicspath{ {./img_dataset/} }
\begin{figure*}[htbp] 
  \centering 
  \includegraphics[width=1.0\textwidth]{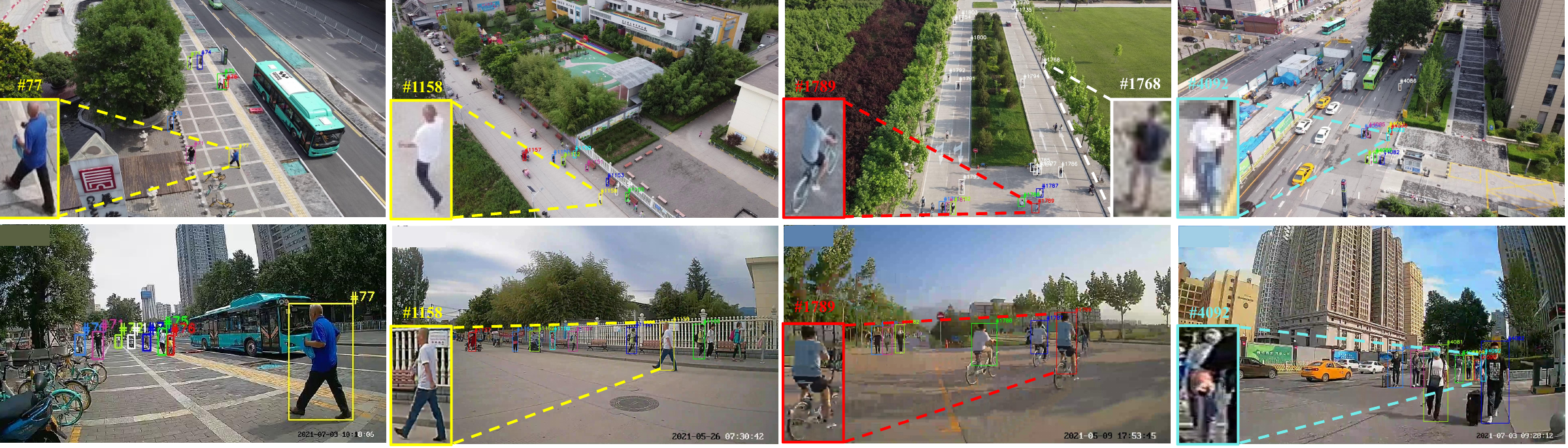} 
  \caption{Exemplars of aerial images and their corresponding ground surveillance images. The first row shows some aerial images, and the second row gives the corresponding ground surveillance images. We show the manually annotated boxes and their IDs for each person, and persons with the same ID from the two views are annotated with the same color. Persons presented in single view are shown with white boxes.} 
  \label{fig:sceneimage} 
\end{figure*}

\section{Related Works}
In this section, we briefly review the related works from the following three aspects:

\textbf{Person Search Datasets.}
As the rapid development of the human-centered visual technology, more and more human-related datasets such as Market1501~\cite{zheng2015scalable}, MSMT17~\cite{wei2018person}, PersonX~\cite{sun2019dissecting}, DukeMTMC-reID~\cite{zheng2017unlabeled} have been collected.  For the person search task, the popular datasets include CUHK-SYSU~\cite{xiao2017joint} and PRW~\cite{zheng2017person}. 
Recent works have achieved very high  performances on them, especially on CUHK-SYSU dataset with mAP of 93.8\%~\cite{li2021sequential}, as these datasets are relatively simple and show small variations in terms of resolution, viewpoint, pose, etc. It is very appealing to collect one large-scale complex dataset from the real-world surveillance scenarios, to promote the development of this research field.


\textbf{Aerial Visual Datasets.}
With the rapid development of the commercial UAVs, many aerial visual datasets~\cite{xia2018dota, cheng2017remote,zhu2018visdrone,zhang2020person,wang2019vehicle, zheng2020university,zhu2023sues} have emerged recently to facilitate the research of aerial visual tasks.
Compared with the traditional visual datasets, these aerial datasets show more challenging intra-class variations in object scale, pose, viewpoint, occlusion, etc.

These tasks mainly focus on the object detection, tracking, crowd counting, while to our knowledge, our G2APS is the first UAV-related dataset for cross-platform person search.
Besides, all these datasets are taken from the traditional single platform of UAVs. In contrast, the constructed G2APS dataset is collected from the cross platform for ground-to-aerial surveillance system, where we need to annotate identities across the UAV and camera, which is more advanced and practical for the real-world surveillance scenarios.



\textbf{Person Search Methods.}
Existing person search methods can be generally divided into two categories: two-step and end-to-end approaches. Usually, the two-step methods~\cite{han2019re,chen2018person,lan2018person,dong2020instance,wang2020tcts,kefeng2021blockchain}  sequentially train a person detector and a person ReID model for person search.
In contrast, the end-to-end person search methods train a unified model for person detection and re-identification for better efficiency.

Usually, the end-to-end methods obtain inferior performance than the two-step approaches, as the jointly learning objectives sometimes conflict with each other and always needs to balance between the detection and re-identification objectives. Therefore, some works~\cite{zhang2019efficient,zhang2021diverse,zhang2021boosting,li2021hierarchical} try to improve the performance of the end-to-end methods by introducing a larger teacher model for knowledge distillation. 

Different from above-mentioned methods, we propose a simple yet efficient knowledge distillation scheme on the head of ReID network, without introducing any extra computation cost during model inference, while with only a very small amount of extra computation cost during model training.

\section{Dataset}
In this section, we firstly devise how to construct and annotate our G2APS  dataset to simulate the Ground-to-Aerial person search task in practice.
Then we mainly highlight the key characteristics of our dataset compared with existing person search datasets.

\begin{figure}[htbp] 
    \centering 
    \includegraphics[width=0.475\textwidth, height=0.35\textwidth]{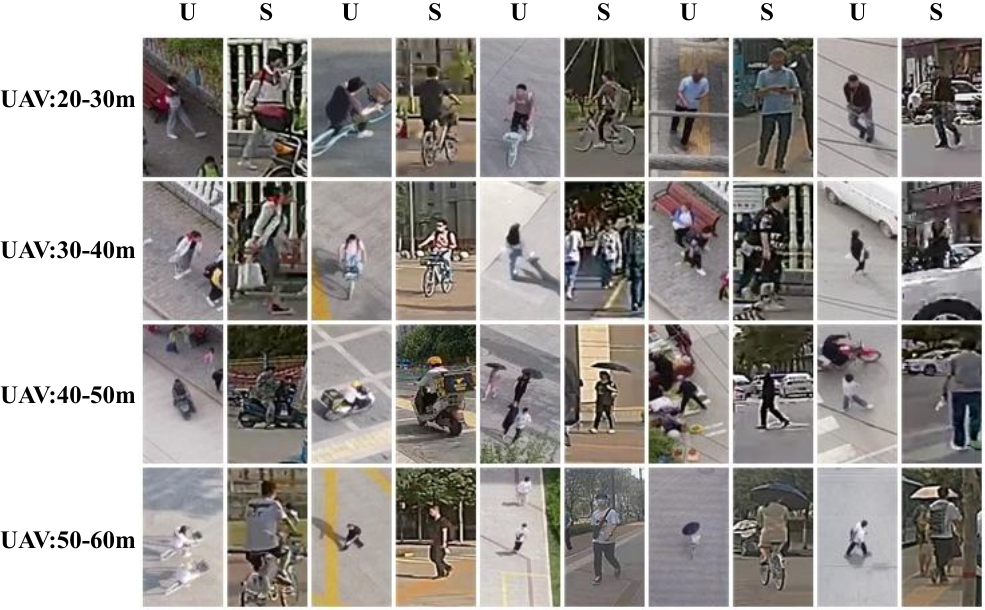} 
    \caption{Visualization and comparison of person images captured by UAV and ground surveillance cameras. The columns denoted as U and S are captured by UAV and ground surveillance camera, respectively. We show the person images captured by UAV with various flying altitudes in each row.} 
    \label{fig:drsu_overview} 
 \vspace{-4mm}
\end{figure}

\subsection{Dataset collection}

During the shooting process, we use a DJI Mavic mini camera and a ground surveillance camera.
The ground camera is fixed about 2 meters above the ground, 
while the camera of the UAV takes pictures at various heights, angles and flight modes in the air, and the flight altitude varies from 20 meters to 60 meters.
In addition, the flight mode of drone includes hovering, cruising and rotating, which makes the captured persons contain richer perspectives.

The paired videos were taken from nine different scenarios, including primary school campus, university campus, subway station entrance, tourist sites, crossroads, sidewalk, and so on.
We collected 36 pairs of videos in total. Then we crop the videos into pictures, leaving 0.5 seconds between the two frames, and it ended up with 31,770 images, with half of them shot by UAV-mounted and ground surveillance camera respectively.
We show some exemplar images taken by the two devices in ~\autoref{fig:sceneimage}.

\subsection{Annotation}

For the annotation of our dataset, we firstly marked the bounding-boxes of all visible persons in the images with the help of a software named Colabeler~\cite{colabeler}.
Through this step, we obtain a total of 260,559 bounding boxes.
Then the identities of the persons are assigned and associated between the paired images captured by UAV and ground surveillance cameras.
Note that the same person is assigned with a unique ID for those persons captured by both the cameras according to the appearance and temporal correspondences, while as for those persons appeared in only one device, their IDs are all denoted as -1.


\begin{figure}[htbp] 
   \centering 
   \includegraphics[width=0.3\textwidth]{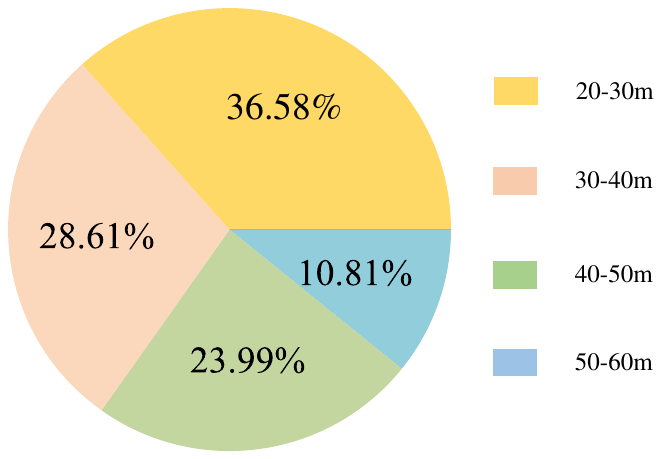} 
   \caption{The distribution over flying altitudes for capturing UAV images.} 
   \label{fig:height_pie} 
\vspace{-4mm}
\end{figure}
        
\begin{figure}[htbp] 
    \centering 
    \includegraphics[width=0.47\textwidth]{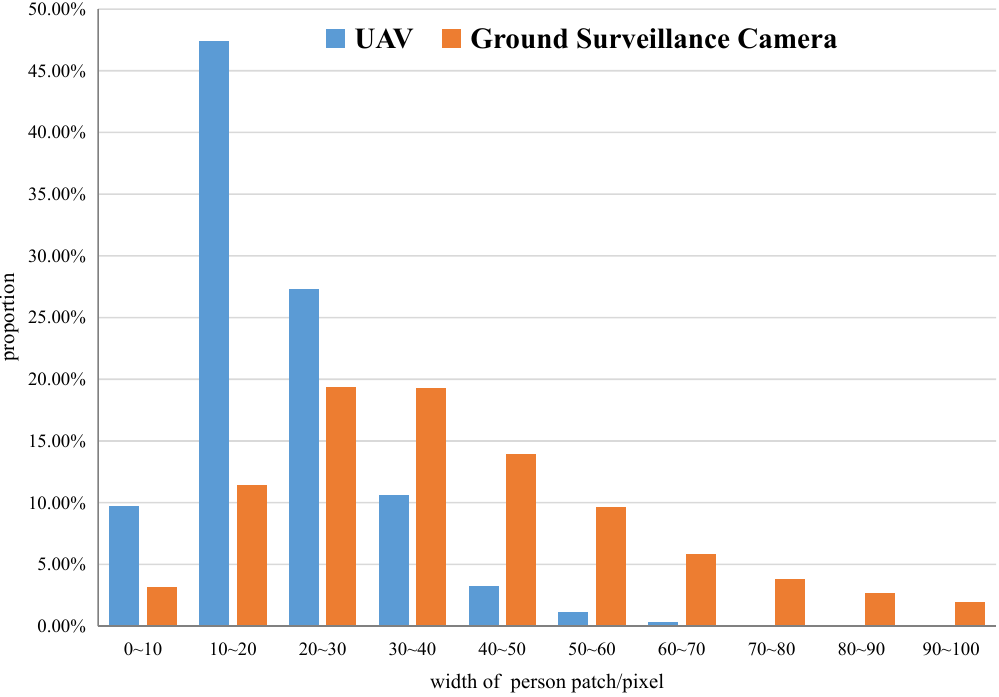} 
    \caption{The distribution over the width of the annotated bounding boxes from ground surveillance camera and UAV, respectively.  } 
    \label{fig:width_dist} 
\vspace{-4mm}
\end{figure}

\subsection{Characteristics of Our G2APS}

Compared with existing popular person search datasets,
our dataset G2APS has the following characteristics:

\textbf{A large amount of labeled data.} 
Our G2APS consists of 2,644 person IDs and 260,559 bounding boxes, of which 199,696 are labeled with unique identities, with an average of 75 bounding boxes per person, much higher than PRW and CUHK-SYSU, as shown in ~\autoref{tab:dataset_compare}. To our knowledge, this is the first large scale ground-to-aerial person search dataset to date.

\textbf{Drastic view changes between query and gallery persons.}
The query and gallery persons are from cross-platform cameras, i.e. ground and aerial views, respectively. Thus the view changes between query and gallery images are drastic compared with existing person search or re-identification datasets.


\textbf{Rich environment scenarios.}
We capture the dataset at multiple locations with rich scenarios, including rural roads, university campus, subway station entrances, tourist sites, sidewalk, and crossroads \textit{etc.}, in order to meet the practical needs in realistic environment of person search.
In contrast, PRW~\cite{zheng2017person} only contains scenes from university campus, while CUHK-SYSU~\cite{xiao2017joint} includes stations, shopping malls and some indoor environments, which are relatively simple.

\textbf{Different resolutions.}
As shown in ~\autoref{fig:height_pie}, the height of UAV-mounted camera varies between a wide range, from 20 to 60 meters, which makes resolutions of the persons very different. 
The width distribution of persons in ground camera captured images is concentrated between 10 and 70 pixels, while that in UAV-captured images is between 5 and 35 pixels. 
~\autoref{fig:width_dist} shows the distribution of person width under both the devices.
Note that the inconsistent resolution distribution between query and gallery images introduces more challenge to the ground-to-aerial person search task.

\textbf{Diverse views and poses.}
Our dataset contains persons with diverse views including profile views and top views, as the flight modes of the UAV includes hovering, cruising and rotating and the mounted camera can be freely adjusted to a large degree.
The ground surveillance camera is fixed on the ground and persons with different poses when walking or riding bicycles are all collected in our dataset.
From ~\autoref{fig:drsu_overview}, it can be seen that there is a huge difference in the perspective and pose for different persons under the UAV and the ground surveillance cameras.

\begin{table}[]
    \caption{Comparison of G2APS with other person search datasets}
    \label{tab:data_statics}
    \begin{NiceTabular}{c|ccc}
    \toprule
    Datasets    & CUHK-SYSU        & PRW    & G2APS            \\ 
    \midrule
    bbox num    & 96,143           & 43,110 & 260,559          \\
    images num  & 18,184           & 11,816 & 31,770           \\
    ID num      & 8,432            & 932    & 2,644            \\
    data source & camera + movie & camera & camera + UAV \\ 
    \bottomrule
    \end{NiceTabular}
\label{tab:dataset_compare}
\end{table}

\begin{figure*}[htbp] 
    \centering 
    \includegraphics[width=\textwidth]{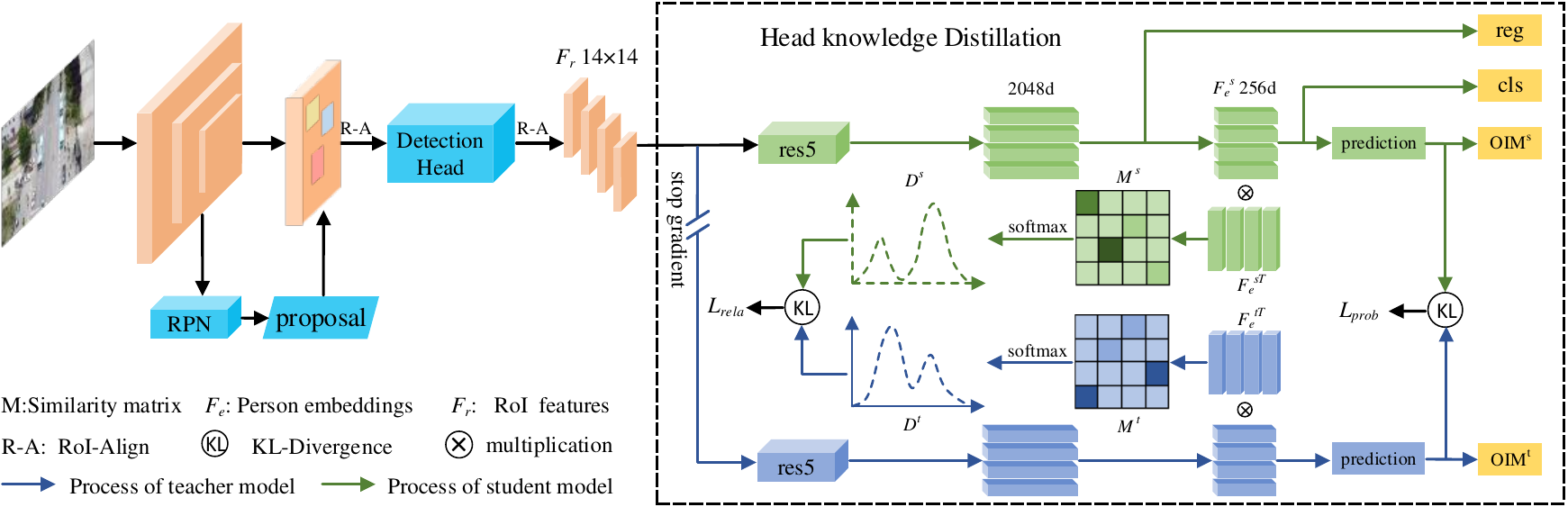} 
    \caption{Overview of our head knowledge distillation(HKD) framework for end-to-end person search.}
    \label{fig:framework} 
\vspace{-4mm}
\end{figure*}

\begin{table}[]
\caption{Performance comparison of end-to-end method and two-step methods}
\resizebox{\linewidth}{!}{
\begin{NiceTabular}{c|c|cc|cc}
\toprule
 Method    & Detector    & Recall & AP    & mAP   & top-1  \\ 
\midrule
two-step   & Faster R-CNN~\cite{ren2015faster} & \textbf{83.60}  & \textbf{68.80} & \textbf{52.58} & \textbf{62.19} \\
two-step   & FCOS~\cite{tian2019fcos}         & 77.40  & 68.10 & 51.86 & 61.84 \\
end-to-end & Faster R-CNN~\cite{ren2015faster} & 74.26  & 66.55 & 40.32 & 50.53 \\ 
\bottomrule
\end{NiceTabular}
}
\label{tab:end_to_end_and_two_step}
\end{table}

\section{Approach}

In this section, we firstly compare existing two-step and end-to-end methods on G2APS and empirically found the bottleneck of end-to-end methods lies in the ReID model. 
Then, to bridge the gap between two-step and end-to-end methods, we propose a Head Knowledge Distillation (HKD) module to alleviate the inconsistence between detection and ReID model and finally improve the ground-to-aerial person search .

\subsection{Bottleneck of the End-to-End Framework}

Generally speaking, two-step models train the object detection and person ReID independently, while end-to-end methods optimize the two tasks jointly.
We evaluate thirteen two-step and seven end-to-end person search methods on our dataset. The best results of the two types of methods are shown in \autoref{tab:end_to_end_and_two_step}, and it can be clearly seen  that two-step methods consistently outperform end-to-end methods by a large margin.


However, to investigate whether the advantage of the two-step model benefitting from excellent detector or stronger ReID model,
we report both the detection and ReID performance on our dataset for the best two-step and end-to-end model respectively.
Note that for two-step method, we adopt both the classical two-stage detector Faster R-CNN~\cite{ren2015faster} and one-stage detector FCOS~\cite{tian2019fcos}, and state-of-the-art HOreid~\cite{wang2020high} is chosen as the ReID model. 
While for end-to-end method we choose the best COAT method~\cite{yu2022cascade}.

It can be seen from the first two rows of ~\autoref{tab:end_to_end_and_two_step} that FCOS achieves inferior recall rate than Faster R-CNN, while the final ReID performance is slightly hampered.
From the second and third rows, it can be seen that although detection performance of end-to-end method is only 1.55\% lower in AP, while the final ReID performance is greatly reduced by 11.54\% in mAP.
Therefore, it can be inferred that for end-to-end model, improving its ReID ability is the key to obtain better ground-to-aerial person search performance.

\subsection{Head knowledge Distillation for End-to-End Person Search}

The performance of the end-to-end method is inferior due to the inconsistent optimization objectives under the joint framework where detection aims to learn features which can distinguish persons from background but ReID aims to learn features which can distinguish persons from each other.
It is especially challenging when it lies great view and pose changes between the query and gallery persons as they are shot by different platform-based cameras.

To alleviate the conflicting objectives, we propose a simple yet effective distillation scheme named Head Knowledge Distillation (HKD) which only introduces an additional ReID head to guide the discriminant feature learning of the whole end-to-end person search method.
The model structure is shown in ~\autoref{fig:framework}.



We set SeqNet~\cite{li2021sequential} as our base model. During training, the proposals predicted by Region Proposal Network (RPN)~\cite{ren2015faster} are first refined to more accurate boxes through the detection head.
Then, RoI-Align is used to pool the boxes into a fixed size to get the RoI feature $F_r$.
$F_r$ is then fed into the ReID head of both the teacher branch and the student branch to extract the feature embeddings for predicting the person IDs. 
The structures of ReID Head in the two branches are devised as the same, both taking the $5_{th}$ stage of ResNet\cite{he2016deep} and being connected with a global average pooling layer and a FC layer to project the features into 256-d embedding vectors.
The teacher branch is trained with only OIM loss~\cite{xiao2017joint} to encourage the features focusing on ReID task. 
In addition, to avoid the interference on backbone detection model, the gradient of the teacher branch would be detached to be not further back-propagated.
Note that the teacher branch is discarded during the inference phase, so our model does not introduce more inference overhead.


\subsection{Training Objectives}


We enforce two types of distillation losses on top of our HKD module, including probability-based knowledge distillation, and relationship-based knowledge distillation.

\textbf{Probability-based knowledge distillation} expects that the student branch can mimic the prediction probability distribution of the teacher branch. 
Specifically, we enforce the KL-Divergence between the probability distributions predicted by the two ReID heads as the probability-based distillation loss $L_{prob}$:

\begin{equation}
    \mathcal L_{prob} =\frac{1}{N} \sum_{i=1}^{N} (KL(p_{i}^{t}||p_{i}^{s})+KL(p_{i}^{s}||p_{i}^{t})),
\end{equation}
where $p_i^t$ and $p_i^s$ denote the predicted probability distribution of the $i_{th}$ sample by the teacher branch and the student branch respectively, and $N$ denotes the total number of persons in the batch. 
KL-Divergence is calculated as follows:
\begin{equation}  \label{eqkl}
    KL(p_{i}^{s}||p_{i}^{t})=\sum _{j=1}^{C}p_{i,j}^{s}log\frac{p_{i,j}^{s}}{p_{i,j}^{t}},
\end{equation}
where $C$ denotes the total number of categories in the training set, and $p_{i,j}$ denotes the probability of $i_{th}$
 sample on $j_{th}$ category.

\textbf{Relation-based knowledge distillation} treats the similarity matrix between samples in a batch as knowledge to guide the student branch to learn the same similarity distribution as the teacher. 
Specifically, we compute the similarity matrixs $M^s,M^t\in \mathbb R^{N\times N}$ among the person embeddings of the two branches respectively:
\begin{equation}
M^{s} =F_e^s\times F_e^{s\top},M^{t} =F_e^t\times F_e^{t\top},
\end{equation}
where $F_e^s$/$F_e^t$ denotes person embeddings extracted by ReID head in the student/teacher branch, and $\top$ means transpose of the matrix.

After softmax normalization processing, the similarity matrixes are converted into probability distributions $D^t,D^s$, and the relationship distillation loss $L_{rela}$ is computed by the KL-Divergence of them:
\begin{equation}
    \mathcal L_{rela}=\frac{1}{N} \sum_{i=1}^{N}KL(D_{i}^{s},D_{i}^{t}). 
\end{equation}


During model training, the loss function of the detector is the same as SeqNet, and the formula is expressed as:
\begin{equation}
\mathcal L_{det}=k_1\mathcal L_{reg1}+k_2\mathcal L_{cls1}+k_3\mathcal L_{reg2}+k_4\mathcal L_{cls2},
\end{equation}\label{detectionLoss}
where $\mathcal L_{reg1}$ and $\mathcal L_{reg2}$ denote the bounding box regression loss on top of the detection head and the ReID head of the student branch respectively, and $\mathcal L_{cls1}$ and $\mathcal L_{cls2}$ represent the classification loss on these two heads accordingly. And $k_1$, $k_2$, $k_3$ and $k_4$ are hyper-parameters to balance each loss.

Additionally, both the student branch and the teacher branch are constrained by OIM loss~\cite{xiao2017joint}, denoted as  $\mathcal L_{oim}^s$ and  $\mathcal L_{oim}^t$.

Online Instance Matching(OIM) Loss~\cite{xiao2017joint} is a popular loss widely used in person search task.It aims to minimize the feature discrepancy among the instances of the same identity, while maximize the discrepancy among persons with different identities.

Finally, the total loss is devised as
\begin{equation}
\mathcal L=\lambda_1 \mathcal{L}_{prob}+\lambda_2 \mathcal L_{rela}+\mathcal L_{det}+L_{oim}^s + \mathcal L_{oim}^t,
\end{equation}\label{OverallLoss}
where $\lambda_1$ and $\lambda_2$ are weight parameters for these two distillation loss $\mathcal L_{_{prob}}$ and $\mathcal L_{rela}$.

\section{Experiments}

\subsection{Evaluation Protocols and Implementation Details}

\textbf{Datasets.}
We conduct experiments on the constructed G2APS dataset and two widely used person search datasets: PRW and CUHK-SYSU. 

\textbf{On G2APS dataset}, there are 2,644 identities with 31,770 images of 260,559 bounding boxes in total. Among them, 2,048 identities with 21,962 images are used for training, and the rest 566 identities with 9,808 images are used for testing.
Specifically, in the testing subset, each identity corresponds to one query image from the ground surveillance camera and 50 gallery images from UAVs. There are 10 images out of the 50 gallery images containing the same identity as the query person image. 
Due to the broad view of UAVs, there will be about 500 persons in the gallery for each query person image, which is quite challenging for person search. The settings of the dataset for training and testing follows the traditional dataset partition in CUHK-SYSU. 

\textbf{CUHK-SYSU} is a large-scale person search dataset composed of 18,184 images with 8,432 identities of 96,143 bounding boxes, from the street snap images and screenshots of films. 
There are  11,206 images of 5,532 identities in the training set, and 2,900 testing identities in the rest 6,978 images with default gallery size as 100.

\textbf{PRW dataset} collects data from six cameras, including 932 identities and 43,110 person bounding boxes in 11,816 images. The training set contains 5,704 images with 482 identities, and the test set includes 6,112 images with 450 identities. For each query, all of the 6,112 images in the test set are set as gallery. ~\autoref{tab:stat_of_pswd} lists more information about all the three datasets.

\begin{table}[]
\caption{Statistics of the G2APS dataset.}
\begin{NiceTabular}{c|cccc}
\toprule
      & \#image & \#ID & \#labeled box & \#unlabeled box \\
\midrule
train & 21,962   & 2,078 & 139,201        & 45,852           \\
test  & 9,808    & 566  & 60,495         & 15,011          \\
\bottomrule
\end{NiceTabular}
\label{tab:stat_of_pswd}
\end{table}

\textbf{Evaluation Protocols.} We follow the standard evaluation metrics for person search~\cite{xiao2017joint,zhong2020robust}. A person is matched if the overlap ratio between the predicted and the ground-truth boxes of the same identity is more than 0.5 intersection over union (IOU). For detection, we adopt Recall and Average Precision(AP) as the evaluation metrics. While for person ReID, the mean Average Precision (mAP) and Cumulative Matching Characteristic (CMC) are adopted as the evaluation metrics.


\textbf{Implementation details.}
We implement our model with PyTorch platform and conduct all experiments on one NVIDIA GeForce RTX 3090 GPU. Following SeqNet~\cite{li2021sequential} ,  ResNet-50~\cite{he2016deep} pretrained on ImageNet is adopted as the backbone. We use the Stochastic Gradient Descent (SGD) optimizer with momentum of 0.9 and the weight decay of 5×$10^{-4}$, to train our model for 21 epochs. 
For G2APS/PRW/CUHK-SYSU datasets, the batch sizes are set to 2/4/3, and the initial learning rates are set to 0.001/0.0018/0.0018, and decreased by a factor of 10 in the $16$-$th$ epoch. 
In addition, the sizes of the circular queue are set as 2000/5000/500 when we compute OIM loss, and the sizes of the lookup table for the three datasets are 2078/5532/482, which is the same as the number of categories $C$ in Eq.~\ref{eqkl}.
The weights for the ReID loss of $\mathcal{L}_{oim}^s$ and $\mathcal{L}_{oim}^t$ are both set to 1.0, and the weight parameters $k_1$, $k_2$, $k_3$ and $k_4$ in the detection loss $\mathcal{L}_{det}$ in Eq.~\ref{detectionLoss} are kept the same with those in the baseline method SeqNet~\cite{li2021sequential}. 
For HKD module, the weight parameters for $\mathcal{L}_{prob}$ and $\mathcal{L}_{rela}$ in Eq.~\ref{OverallLoss} are set as $\lambda_1$=1.0 and $\lambda_2$=300, respectively.


\subsection{Comprehensive Evaluation on G2APS Dataset.}
We comprehensively evaluate both two-step and end-to-end person search methods on our dataset.
 
\textbf{Two-step Methods.}
Two step methods divide the person search task into person detection and ReID tasks.
We adopt two representative detectors Faster R-CNN~\cite{ren2015faster} and FCOS~\cite{tian2019fcos}, and choose thirteen popular person ReID methods to comprehensively evaluate the ground-to-aerial person search task on G2APS.

Note that for two step methods, we firstly train person detection models based on the bounding box annotations. 
Then, person ReID models are trained based on the cropped person patches predicted by the detector.
During inference, for fair comparison with end-to-end methods, all person patches detected from those 50 gallery images are treats as candidates for each query person.
The detection performances are reported in ~\autoref{tab:end_to_end_and_two_step}. 
It can be seen that the two-stage Faster R-CNN detector~\cite{ren2015faster} achieves better detection results with 6.2\% higher recall rate and 0.7\% AP gains compared with the one-stage FCOS detector~\cite{tian2019fcos}.
We report the final person search results in ~\autoref{tab:twostep_ps}.

\begin{table}[]
\caption{Performance comparison of two-step person search methods based on Faster R-CNN and FCOS combined with other ReID models on Our Dataset.}
\begin{NiceTabular}{l|ll|ll}
\toprule
\multirow{2}{*}{Method}        & \multicolumn{2}{c}{Faster R-CNN} & \multicolumn{2}{c}{FCOS} \\ \cline{2-5}
              & mAP           & rank-1         & mAP        & rank-1      \\
\midrule
HOreid~\cite{wang2020high}        & \textbf{52.58}         & \textbf{62.19}          & \textbf{51.86}      & \textbf{61.84}       \\
LUPnl~\cite{fu2022large}     & 50.24         & 61.31          & 49.84      & 61.84       \\
CDNET~\cite{li2021combined}         & 49.53         & 58.13          & 50.53      & 57.24       \\
Bag-of-Tricks~\cite{luo2019bag} & 48.49         & 55.83          & 47.54      & 57.77       \\
PFD~\cite{wang2022pose}           & 48.01         & 55.83          & 47.33      & 54.06       \\
GASM~\cite{he2020guided}          & 46.82         & 55.12          & 46.6       & 57.95       \\
Align++~\cite{luo2019alignedreid++}       & 46.36         & 55.65          & 45.35      & 54.77       \\
Unreal~\cite{zhang2021unrealperson}        & 46.15         & 55.48          & 45.7       & 56.36       \\
DG-Net~\cite{zheng2019joint}        & 44.66         & 54.59          & 44.02      & 54.06       \\
CBN~\cite{zhuang2020rethinking}           & 44.58         & 53.53          & 43.86      & 55.65       \\
PCB~\cite{sun2018beyond}           & 43.95         & 52.65          & 44.35      & 51.77       \\
CtF~\cite{wang2020faster}           & 43.51         & 53.89          & 42.86      & 53.36       \\
MGN~\cite{wang2018learning}           & 39.42         & 46.82          & 38.88      & 47.88       \\
\bottomrule
\end{NiceTabular}
\label{tab:twostep_ps}
\end{table}





As can be seen from ~\autoref{tab:twostep_ps},
among these methods, the one which adopts HOreid~\cite{wang2020high} as ReID model and Faster RCNN as detector has achieved the best results. 
The possible reason is that, when the view differences between the ground surveillance image and the UAV image is large, the flexible matching process based on the graph topology proposed by HOreid can better illustrate the corresponding human body parts between these two images. 
In addition, CDNET~\cite{li2021combined} fuses RGB and depth image information to improve the feature representation ability of the network. 
LUPnl~\cite{fu2022large} obtains a more robust backbone network through pre-training on a large-scale noisy data. 
Bag-of-Tricks~\cite{luo2019bag} uses a series of simple and effective training tricks to construct a powerful baseline model. 
The pose-guided feature decoupling strategy proposed in PFD~\cite{wang2022pose} effectively alleviates the negative effects of object occlusion. 
Therefore, they all achieve better person search results relatively.


However, since the person images captured by the UAV are relatively small, and there always contains severe self-occlusion from an aerial view, it is difficult for the methods based on part features such as Align++~\cite{luo2019alignedreid++}, PCB~\cite{sun2018beyond}, and MGN~\cite{wang2018learning} to accurately match the stripe areas of two persons under the perspective of surveillance camera and UAV. As a result, their performances are inferior compared with other methods. 
All these methods on top of different person detectors demonstrate similar performance characteristics, as shown in ~\autoref{tab:twostep_ps}.

\textbf{End-to-End Methods.}
Besides two-step methods, we also conduct experiments on G2APS with seven representative end-to-end person search methods, where the experimental results are reported in Table~\ref{tab:sota_G2APS}.
It can be seen that end-to-end methods generally achieve inferior results compared with the two-step approaches.
Although COAT~\cite{yu2022cascade} achieves the best results among these end-to-end methods, it is still inferior to the best two-step method HOreid+Faster R-CNN~\cite{wang2020high} by a large margin of 12.3\% mAP, which indicates that the inconsistency training objective between detection and ReID is especially unavoidable for ground-to-aerial person search task. Table~\ref{tab:end_to_end_and_two_step} shows that the bottleneck of end-to-end methods lies in the ReID model, which motivates us to propose the HKD mechanism to alleviate the conflicting objectives in such end-to-end person search methods.



\begin{table}[]
\caption{Performance Evaluation of End-to-End Person Search Methods on Our G2APS Dataset.}
\begin{NiceTabular}{l|ll}
\toprule
\multirow{2}{*}{Method}     & \multicolumn{2}{c}{G2APS}     \\ \cline{2-3}
\multicolumn{1}{c}{}           & mAP          & top-1        \\
\midrule
Faster R-CNN~\cite{ren2015faster}+HOreid~\cite{wang2020high}                       & {52.58}                   & {62.19} \\
FCOS~\cite{tian2019fcos}+HOreid~\cite{wang2020high}  & {51.86}                   & {61.84} \\
\midrule
OIM~\cite{xiao2017joint}          & 31.16              & 38.52    \\
NAE~\cite{chen2020norm}          & 30.95                     & 39.22                     \\
AlignPS~\cite{yan2021anchor}      & 26.99                     & 34.68                     \\
OIM++~\cite{lee2022oimnet++}        & 32.5                      & 40.28                     \\
SeqNet~\cite{li2021sequential}       & 33.96                     & 44.52                     \\
PSTR~\cite{cao2022pstr}         & 28.36                     & 39.93                     \\
COAT~\cite{yu2022cascade}         & 40.32                     & 50.53                     \\
\midrule            
SeqNet+HKD   & 39.40(+5.44)              & 49.12(+4.60)               \\
COAT+HKD     & \textbf{41.41}(+1.09)              & \textbf{51.94}(+1.41)              \\
\bottomrule
\end{NiceTabular}
\label{tab:sota_G2APS}
\end{table}

\subsection{Comparison with State-of-the-Art Methods}

To bridge the gap between end-to-end methods and two-step methods, we propose the ReID network head based knowledge distillation mechanism, on top of two representative end-to-end person search methods, i.e., SeqNet~\cite{li2021sequential} and COAT~\cite{yu2022cascade}. The experimental results are shown in ~\autoref{tab:sota_G2APS} and~\autoref{tab:sota_prw}. It can be clearly seen that with the help of HKD, the performances of SeqNet+HKD get improved by a large margin of 5.44\%mAP and 4.6\% rank-1 accuracy on G2APS dataset, compared with the baseline method SeqNet~\cite{li2021sequential}. While for state-of-the-art end-to-end method COAT~\cite{yu2022cascade}, the performances can still be improved with 1.09\% mAP gains and 1.39\% gains in rank-1 accuracy. 



Additionally, to further show the effectiveness of our proposed HKD mechanism, we also conduct experiments on two widely adopted PRW and CUHU-SYSU datasets and report the experimental results in ~\autoref{tab:sota_prw}. On CUHK-SYSU dataset, HKD improves SeqNet with 1.45\% mAP gains and 1.5\% gains in rank-1 accuracy. Meanwhile, HKD also improves COAT with 0.18\% mAP gains and 0.66\% rank-1 accuracy gains. On PRW dataset, HKD improves SeqNet with 4.79\% mAP gains and 1.72\% gains in rank-1 accuracy, and HKD improves COAT with 1.04\% mAP gains and 0.63\% rank-1 accuracy gains. Finally, it is worth noting that the proposed HKD mechanism takes current approaches to a new state-of-the-art on all the three datasets.

\begin{table}[]\small
\caption{ Comparison with the state-of-the-art methods on PRW and CUHK-SYSU. * indicates that the results are implemented by ourselves with the open source code.}
\resizebox{\linewidth}{!}{
\begin{NiceTabular}{l|ll|ll}
\toprule
\multirow{2}{*}{Method}  & \multicolumn{2}{c}{PRW} & \multicolumn{2}{c}{CUHK-SYSU} \\\cline{2-5}
                      & mAP           & top-1         & mAP        & top-1      \\
\midrule
DPM~\cite{zheng2017person}               & 20.5       & 48.3        & -             & -               \\
MGTS~\cite{chen2018person}                & 32.6       & 72.1     & 83.0          & 83.7              \\
CLSA~\cite{lan2018person}                   & 38.7       & 65.0  	& 87.2          & 88.5            \\
RDLR~\cite{han2019re}                    & 42.9       & 70.2       & 93.0          & 94.2             \\
IGPN~\cite{dong2020instance}             & 47.2       & 87.0        & 90.3          & 91.4            \\
TCTS~\cite{wang2020tcts}                 & 46.8       & \textbf{87.5}   	& 93.9          & 95.1               \\
\hline
OIM~\cite{xiao2017joint}                 & 21.3       & 49.9     	& 75.5          & 78.7            \\
IAN~\cite{xiao2019ian}                	 & 23.0       & 61.9     	& 76.3          & 80.1            \\
NPSM~\cite{liu2017neural}                & 24.2       & 53.1        	& 77.9          & 81.2           \\
CTXG~\cite{yan2019learning}              & 33.4       & 73.6        & 84.1          & 86.5            \\
QEEPS~\cite{munjal2019query}             & 37.1       & 76.7         & 88.9          & 89.1           \\
HOIM~\cite{chen2020hierarchical}         & 39.8       & 80.4         & 89.7          & 90.8           \\
APNet~\cite{zhong2020robust}             & 41.9       & 81.4         & 88.9          & 89.3           \\
BINet~\cite{dong2020bi}                 & 45.3       & 81.7      	& 90.0          & 90.7           \\
NAE~\cite{chen2020norm}                 & 43.3       & 80.9       	& 91.5          & 92.4            \\
DMRNet~\cite{han2021decoupled}           & 46.9       & 83.3         & 93.2          & 94.2           \\
PGS~\cite{kim2021prototype}              & 44.2       & 85.2         & 92.3          & 94.7           \\
AlignPS~\cite{yan2021anchor}             & 45.9       & 81.9         & 93.1          & 93.4           \\
DMRNet++~\cite{han2022dmrnet++}            & 51.0       & 86.8         & 94.4          & 95.5           \\
SeqNeXt+GFN~\cite{jaffe2023gallery}            & 51.3       & 90.6         & 94.7          & 95.3           \\
SeqNet~\cite{li2021sequential}           & 46.7       & 83.4         & 93.8          & 94.6           \\
COAT* ~\cite{yu2022cascade}              & 52.45      & 86.00        & 93.68 		& 94.10  		   \\
\midrule
SeqNet+HKD  		& 51.49(+4.79) & 85.12(+1.72)  & \textbf{95.25}(+1.45)   & \textbf{96.10}(+1.5)     \\
COAT*+HKD   		& \textbf{53.49}(+1.04) & 86.63(+0.63)   & 93.86(+0.18) 			& 94.76(+0.66)        \\
\bottomrule
\end{NiceTabular}
}
\label{tab:sota_prw}
\end{table}

\subsection{Ablation Study}

\textbf{Effectiveness of HKD.}
The proposed HKD module contains two novel ingredients: the probability-based and relation-based knowledge distillation components. To reveal how each ingredient contributes to the performance improvement, we conduct ablation study on the G2APS dataset with these two types of distillation losses, and the experimental results are shown in Table~\ref{tab:HKD}.
When only adding HKD to the baseline SeqNet model~\cite{li2021sequential} and using no additional distillation losses, the mAP gets improved from 33.96\% to 34.19\%, which indicates that the newly added teacher branch has negligible impact on the final model performance. 
After enforcing $\mathcal L_{prob}$ or $\mathcal L_{rela}$ on HKD, the performance can be increased from 33.96\% mAP to 39.02\% mAP or 37.70\% mAP, respectively. When the two losses are applied simultaneously, the model performance finally reaches to 39.40\% mAP, showing the effectiveness of the proposed HKD mechanism.

\begin{table}[]
\caption{Effectiveness of the proposed HKD with two distillation losses on the G2APS. $\checkmark$ means applying corresponding loss to HKD, while $\times$  means training model without using it.}
\begin{NiceTabular}{l|cc|cc}
\toprule
        & $\mathcal{L}_{kdp}$  & $\mathcal{L}_{rela}$ & mAP   & top-1 \\
\midrule
two-step & -      & -    & 52.58 & 62.19 \\
\midrule
SeqNet     & -      & -    & 33.96 & 44.52             \\
SeqNet+HKD & $\times$      & $\times$    & 34.19 & 42.40           \\
SeqNet+HKD &        & \checkmark    & 37.70 & 47.70         \\
SeqNet+HKD & \checkmark      &      & 39.02 & 48.06           \\
SeqNet+HKD & \checkmark      & \checkmark    & \bf{39.40} & \bf{49.12}          \\
\bottomrule
\end{NiceTabular}
\label{tab:HKD}
\end{table}


\textbf{Whether or Not Detach the Gradient of Teacher Branch.}
For the teacher branch, we detach the back-propagated gradient flow from the loss function in the teacher branch to the detection network, to avoid the interference of the detection module.
We conduct experiments to verify whether or not detaching the gradient of the teacher branch, and the results are shown in Table~\ref{tab:detachgrad}.
On the basis of the knowledge distillation with only $\mathcal{L}_{prob}$ loss, when we train the model without using the detach technique, the ReID performance will be decreased from 39.02\% mAP to 38.19\% mAP, and detection performance drops from 67.81\%AP to 64.67\%AP. This indicates that detaching the gradient of teacher branch will further alleviate the training conflicting problem between detection and ReID tasks.


\begin{table}[]
\caption{Comparison of model performance with/without detaching the gradient of the teacher branch.}
\vspace{-2mm}
\begin{NiceTabular}{l|c|cc|cc}
\toprule
         & detach & mAP   & top-1 & Recall & AP    \\
        \midrule
SeqNet+HKD     &  \checkmark              & \textbf{39.02} & \textbf{48.06} & \textbf{74.10}  & \textbf{67.81} \\
SeqNet+HKD     & $\times$              & 38.19 & 47.53 & 71.55  & 64.67 \\
\bottomrule
\end{NiceTabular}
\label{tab:detachgrad}
\end{table}

\section{Conclusion}
In this paper, we are the first to construct a large-scale ground-to-aerial person search benchmark dataset, named G2APS, for the cross-platform ground-to-aerial intelligent surveillance applications. The dataset consists of 31,770 images of 260,559 annotated bounding boxes for 2,644 identities.
Comprehensive experiments are conducted on this dataset with 13 two-step and 7 end-to-end person search methods. Besides, we also propose a Head Knowledge Distillation module to alleviate the conflicting training objectives by introducing an additional teacher branch for ReID.
We hope our work can contribute to the development of the researches on the cross-platform ground-to-aerial person search task.



\begin{acks}
This work was supported in part by the National Natural Science Foundation of China (NSFC) under Grant 62101453, Grant U19B2037,  Grant 62176198  and Grant 62201467; in part by the Guangdong Basic and Applied Basic Research Foundation under Grant 2021A1515110544; in part by the Natural Science Basic Research Program of Shaanxi under Grant 2022JQ-686, 2019JQ-158, and in part by the Project funded by China Postdoctoral Science Foundation under Grant 2022TQ0260, and in part by the Young Talent Fund of Xi'an Association for Science and Technology under Grant 959202313088.
\end{acks}

\balance
\bibliographystyle{ACM-Reference-Format}
\bibliography{sample-base}

\appendix
\section{Appendix}

\textbf{Distance Measurements for Sample Relationship Matrices.}
To extensively analyze the proposed HKD module, we implement three widely used distance measurements on the knowledge distillation framework: KL-Divergence, Mean Square Error(MSE)~\cite{tung2019similarity} and Mutual Information~\cite{passalis2018learning}. The comparison results are shown in Table~\ref{tab:relation_metrics}. 
Among the three methods, KL-Divergence achieves the best results. 
The reason is that KL-Divergence is more suitable for describing the distance between two distributions.

\begin{table}[]
\caption{Performance comparison of distance measurements for sample relationship matrices in HKD.}
\vspace{-2mm}
\begin{NiceTabular}{l|cc}
\toprule
                   & mAP   & top-1 \\
\midrule
KL-Divergence                 & \textbf{37.70} & \textbf{47.70} \\
Mutual Information~\cite{passalis2018learning} & 35.88 & 44.70 \\
MSE~\cite{tung2019similarity}                & 35.30 & 43.46 \\
\bottomrule
\end{NiceTabular}
\label{tab:relation_metrics}
\end{table}

\textbf{Evaluation in images captured from various heights.}In the task of ground-to-aerial person search, the ability to retrieve UAV images captured from various flying altitudes is an important basis to measure the performance of the model. Therefore, we select query and gallery set of different heights from the test set to form four different subsets. Then all end-to-end methods are evaluated on these subsets, and results are shown in ~\autoref{tab: test_differ_height}. It can be seen that the resolution of pedestrian image gradually decrease with the increase of camera height, resulting in pedestrian matching task becoming more and more difficult. In the test subset with camera height from 20 to 30 meters, the performance can reach to 55.87\% mAP, while in the test subset with camera height from 50 to 60 meters, the performance can only reach to 18.72\% mAP. 

Obviously, the evaluation results on all the subsets demonstrate that the proposed method HKD helps to boost the baseline method by a large margin, and finally we obtain superior performances to the compared methods on all the experiment settings consistently.


\begin{table*}[]
\caption{Performance of all end-to-end methods on aerial-view images captured from various heights.}
\begin{NiceTabular}{l|ll|ll|ll|ll|ll}
\toprule
\multirow{2}{*}{Method}    & \multicolumn{2}{c}{20-30m}      & \multicolumn{2}{c}{30-40m}      & \multicolumn{2}{c}{40-50m}      & \multicolumn{2}{c}{50-60m}      & \multicolumn{2}{c}{full test dataset} \\ \cline{2-11}
           & mAP            & top-1          & mAP            & top-1          & mAP            & top-1          & mAP            & top-1          & mAP               & top-1             \\
\midrule
OIM~\cite{xiao2017joint}        & 41.26          & 47.60           & 32.05          & 44.57          & 20.36          & 25.98          & 15.31          & 14.29          & 31.16             & 38.52             \\
NAE~\cite{chen2020norm}        & 44.40           & 55.77          & 31.47          & 42.29          & 17.15          & 23.62          & 11.99          & 10.71          & 30.95             & 39.22             \\
AlignPS~\cite{yan2021anchor}    & 33.04          & 38.94          & 25.85          & 32.00             & 23.43          & 36.22          & 16.20           & 19.64          & 26.99             & 34.28             \\
OIM++~\cite{lee2022oimnet++}      & 42.55          & 52.40           & 36.46          & 44.01          & 18.53          & 22.05          & 12.27          & 16.07          & 32.50              & 40.28             \\
SeqNet~\cite{li2021sequential}     & 47.04          & 58.65          & 35.72          & 48.01          & 19.90           & 30.71          & 11.81          & 12.50           & 33.96             & 44.52             \\
PSTR~\cite{cao2022pstr}       & 39.30           & 63.46          & 27.52          & 41.14          & 17.42          & 33.07          & 15.93          & 19.64          & 28.36             & 39.93             \\
COAT~\cite{yu2022cascade}       & 54.24          & 66.83          & 43.58          & 57.14          & 23.29          & 33.86          & 17.09          & 21.43          & 40.32             & 50.53             \\
SeqNet+HKD & 51.64          & 61.06          & 42.97          & 56.57          & 24.09          & 32.28          & 17.50           & 19.64          & 39.40              & 49.12             \\
COAT+HKD   & \textbf{55.87} & \textbf{66.35} & \textbf{43.71} & \textbf{56.57} & \textbf{24.55} & \textbf{35.43} & \textbf{18.72} & \textbf{21.43} & \textbf{41.41}    & \textbf{51.94}   \\
\bottomrule
\end{NiceTabular}
\label{tab: test_differ_height}
\end{table*}

\end{document}